\def\year{2019}\relax
\documentclass[letterpaper]{article} 
\usepackage{aaai19}  
\usepackage{times}  
\usepackage{helvet}  
\usepackage{courier}  
\usepackage{url}  
\usepackage{graphicx}  
\usepackage{url}            
\usepackage{amsmath}
\usepackage{booktabs}       
\usepackage{amsfonts}       
\usepackage{nicefrac}       
\usepackage{microtype}      

\usepackage{pgfplots,capt-of}
\usepackage{accents}
\newcommand{\vectr}[1]{\accentset{\rightarrow}{\mathbf{#1}}}
\newcommand{\vectl}[1]{\accentset{\leftarrow}{\mathbf{#1}}}
\usepackage{booktabs}
\usepackage{multirow}
\usepackage{caption} 
\newcommand{\STAB}[1]{\begin{tabular}{@{}c@{}}#1\end{tabular}}

\begin{document}
%
\title{Improved Sentence Modeling using Suffix Bidirectional LSTM}
\author{ Siddhartha Brahma\\
IBM Research AI, Almaden, USA
}
\maketitle

\begin{abstract}
Recurrent neural networks have become ubiquitous in computing representations of sequential data, especially textual data in natural language processing. In particular, Bidirectional LSTMs are at the heart of several neural models achieving state-of-the-art performance in a wide variety of tasks in NLP. However, BiLSTMs are known to suffer from sequential bias -- the contextual representation of a token is heavily influenced by tokens close to it in a sentence. We propose a general and effective improvement to the BiLSTM model which encodes each suffix and prefix of a sequence of tokens in both forward and reverse directions. We call our model Suffix Bidirectional LSTM or \textbf{SuBiLSTM}. This introduces an alternate bias that favors long range dependencies. We apply SuBiLSTMs to several tasks that require sentence modeling. We demonstrate that using SuBiLSTM instead of a BiLSTM in existing models leads to improvements in performance in learning general sentence representations, text classification, textual entailment and paraphrase detection. Using SuBiLSTM we achieve new state-of-the-art results for fine-grained sentiment classification and question classification.
\end{abstract}

\section{Introduction}
Recurrent Neural Networks (RNN) \cite{Elman1990} have emerged as a powerful tool for modeling sequential data. Vanilla RNNs have largely given way to more sophisticated recurrent architectures like Long Short-Term Memory \cite{Hochreiter1997} and the simpler Gated Recurrent Unit \cite{Cho2014}, owing to their superior gradient propagation properties. 
The importance of LSTMs in natural language processing, where a sentence as a sequence of  tokens represents a fundamental unit, has risen exponentially over the past few years. A LSTM processing a sentence  in the forward direction produces distributed representations of its prefixes. 
A Bidirectional LSTM (BiLSTM in short) \cite{Schuster1997}\cite{Graves2005} additionally processes the sentence in the reverse direction (starting from the last token) producing  representations of the suffixes (in the reverse direction). For every token $t$ in the sentence, a BiLSTM thus produces a contextual representation of $t$ based on its prefix and suffix in the sentence.


Despite their sophisticated design, it is well known that LSTMs  suffer from sequential bias \cite{Pascanu2013}. The hidden state of a LSTM is heavily influenced by the last few tokens it has processed. This implies that the contextual representation of $t$ is highly influenced by the tokens close to it in the sequential order, with tokens farther away being less influential. Computing contextual representations that capture long range dependencies is a challenging research problem, with numerous applications. 

In this paper, we propose a simple, general and effective technique to compute contextual representations that capture long range dependencies.
For each token $t$, we encode both its prefix and suffix in both the forward and reverse direction. Notably, the encoding of the suffix in the forward direction is biased towards tokens sequentially farther away to the right of $t$. Similarly, the encoding of the prefix in the reverse direction is biased towards tokens sequentially farther away to the left of $t$.  Further, we combine the prefix and suffix representations by a simple max-pooling operation to produce a richer contextual representation of $t$ in both the forward and reverse direction. 
We call our model Suffix BiLSTM or \textbf{SuBiLSTM} in short.   A SuBiLSTM has the same representation length as a BiLSTM with the same hidden dimension. 
 
 \begin{figure*}
\centering
\includegraphics[width=0.8\linewidth]{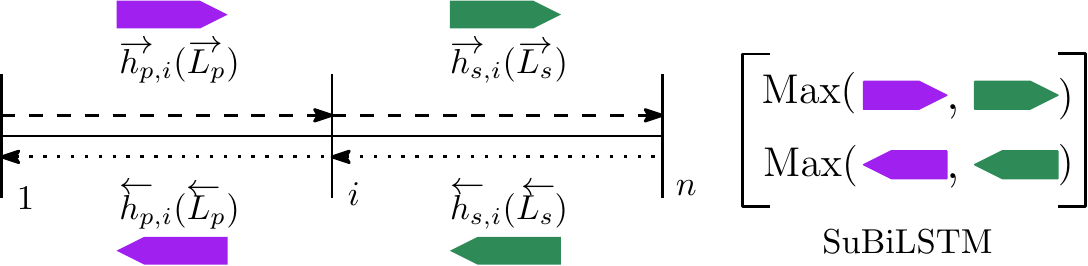}
\caption{Schematics of SuBiLSTM. The large solid purple arrow represents prefixes and large solid seagreen arrow represents suffixes. Their directions represent the encoding direction of the corresponding LSTMs. Best viewed in color.}
\label{fig:schematic}
\end{figure*}
We consider two versions of SuBiLSTMs -- a \emph{tied} version where the suffixes and prefixes in each direction are encoded using the same LSTM and an untied version where two different LSTMs are used. Note that, as in a BiLSTM, we always use different LSTMs for the forward and reverse direction.
In general a SuBiLSTM  can be used as a drop in replacement in any model that uses the intermediate states of a BiLSTM, without changing any other parts of the model. However, the main motivation for introducing SuBiLSTMs is to apply it to problems that require whole sentence modeling e.g. text classification, where the richer contextual information can be helpful. We demonstrate the effectiveness of SuBiLSTM on several sentence modeling tasks in NLP -- general sentence representation, text classification, textual entailment and paraphrase detection. In each of these tasks, we show gains by simply replacing BiLSTMs in strong base models, achieving a new state-of-the-art in fine grained sentiment classification and question classification. 

\section{Suffix Bidirectional LSTM}
Let $\mathbf{s}$ be a sequence with $n$ tokens. We use $\mathbf{s}[i\colon j]$ to denote the sequence of embeddings of the tokens from $\mathbf{s}[i]$ to $\mathbf{s}[j]$, where $j$ maybe less than $i$. 
Let $\vectr{L}_p$ represent a LSTM that encodes \textit{prefixes} of $\mathbf{s}$ in the \emph{forward} direction. 
For the $i$-th token $\mathbf{s}[i]$, we have
\begin{equation}
\vectr{h}_{p,i} = \vectr{L}_p(\mathbf{s}[1\colon i])
\end{equation}
Let $\vectr{L}_s$ represent a LSTM that encodes \emph{suffixes} of $\mathbf{s}$ in the \emph{forward} direction. 
\begin{equation}
\vectr{h}_{s,i} = \vectr{L}_s(\mathbf{s}[i\colon n])
\end{equation}
Note that the $\vectr{h}_{p,i} $ can be computed in a single pass over $\mathbf{s}$, while computing $\vectr{h}_{s,i}$ needs a total of $n$ passes over progressively smaller suffixes of $\mathbf{s}$. 
Now consider $\vectl{L}_p$ and $\vectl{L}_s$ that encodes the prefixes and suffixes of $\mathbf{s}$  in the \emph{reverse}  direction. 
\begin{eqnarray}
&\vectl{h}_{p,i} = \vectl{L}_p(\mathbf{s}[i\colon 1]) \\
&\vectl{h}_{s,i} = \vectl{L}_s(\mathbf{s}[n\colon i]) 
\end{eqnarray}
Note that both $\vectr{h}_{p,i}$ and $\vectl{h}_{p,i}$ encode the same prefix, but in different directions. Similarly,  $\vectr{h}_{s,i}$ and $\vectl{h}_{s,i} $ encode the same suffix, but in different directions. See Fig. \ref{fig:schematic} for a schematic illustration. 

We have four vectors  $\vectr{h}_{p,i}, \vectr{h}_{s,i}, \vectl{h}_{p,i}, \vectl{h}_{s,i}$ that constitute the context of $\mathbf{s}[i]$. Using these, we define the following contextual representation of $\mathbf{s}[i]$.
\begin{eqnarray}
\mathbf{H}_{i}^{\text{SuBiLSTM}} = \left[\max\left\{\vectr{h}_{p,i}, \vectr{h}_{s,i}\right\}; \max\left\{\vectl{h}_{p,i}, \vectl{h}_{s,i}\right\}\right]
\label{eq:subidef}
\end{eqnarray}
Here \textbf{;} is the concatenation operator. This defines the  \textbf{SuBiLSTM} model. We also define another representation where the two LSTMs encoding the sequence in the \emph{same}  direction are the same or their weights are tied. This defines the \textbf{SuBiLSTM-Tied} model, which concretely is 
\begin{eqnarray}
\mathbf{H}_{i}^{\text{SuBiLSTM-Tied}}  &= \left[\max\left\{\vectr{h}_{p,i}, \vectr{h}_{s,i}\right\}; \max\left\{\vectl{h}_{p,i}, \vectl{h}_{s,i}\right\}\right] \\
  &\text{ where } \vectr{L}_p\equiv \vectr{L}_s, \vectl{L}_p\equiv \vectl{L}_s  \notag
\end{eqnarray}
In contrast to SuBiLSTM, a standard BiLSTM uses the following contextual representation of $\mathbf{s}[i]$. 
\begin{eqnarray}
\mathbf{H}_{i}^{\text{BiLSTM}}  = \left[\vectr{h}_{p,i}; \vectl{h}_{s,i}\right] 
\end{eqnarray}
For a fixed hidden dimension, SuBiLSTM and SuBiLSTM-Tied have the same representation length as a  BiLSTM. Importantly,  SuBiLSTM-Tied uses the same number of parameters as a BiLSTM,  while SuBiLSTM uses twice as many.

\subsection{Interpretations of SuBiLSTM }
Notice that $\vectr{h}_{s,i}$ is biased towards tokens that are sequentially to the right and farthest away from $\mathbf{s}[i]$. Combining it with $\vectr{h}_{p,i}$ which is influenced more by tokens close to and to the left of $\mathbf{s}[i]$  creates a representation of $\mathbf{s}[i]$ that is dependent on and influenced by tokens both close and far away from it. The same argument can be repeated in the reverse direction with $\vectl{h}_{p,i}$ and $\vectl{h}_{s,i}$. We argue that this is a richer contextual representation of $\mathbf{s}[i]$ which can help in better sentence modeling, as compared to BiLSTMs where the representation is biased towards sequentially close tokens.

As an alternate viewpoint, for every token  $\mathbf{s}[i]$, SuBiLSTM creates two representations of its prefix  $\mathbf{s}[1\colon i]$,  $\vectr{h}_{p,i}$ and $\vectl{h}_{p,i}$. Their concatenation $[\vectr{h}_{p,i}; \vectl{h}_{p,i}]$ is equivalent to an encoding of the prefix with a BiLSTM consisting of $\vectr{L}_p$ and $\vectl{L}_p$. Similarly, $[\vectr{h}_{s,i}$; $\vectl{h}_{s,i}]$ is an encoding of the suffix $\mathbf{s}[i\colon n]$ by a BiLSTM consisting of  $\vectr{L}_s$ and $\vectl{L}_s$. Thus  $\mathbf{H}_{i}^{\text{SuBiLSTM}} $ can be interpreted as the max-pooling of the bidirectional representations of the prefix and suffix of  $\mathbf{s}[i]$ into a compact representation. This may be contrasted with a BiLSTM where the prefix is encoded by a  LSTM in the forward direction and the suffix is encoded  by another LSTM in the reverse direction. SuBiLSTM thus tries to capture more information by encoding the prefix and suffix in a bidirectional manner. 

In general, the prefix and suffix encodings can be combined in other ways e.g. concatenation, mean or through a learned gating function. However, we use max-pooling  because it is a simple parameterless operation and it performs better in our experiments. Since both SuBiLSTM and SuBiLSTM-Tied produces representations of each token $\mathbf{s}[i]$ in the same way as a BiLSTM, they can be used as drop in replacements for a BiLSTM in any model that uses these  representations. 

\subsection{Time complexity of a SuBiLSTM}
To compute the contextual representations of a minibatch of sentences using a SuBiLSTM, we calculate all the $\vectr{h}_{p,i}$ in one pass using $\vectr{L}_p$. We then create several minibatches (determined by the maximum length of a sentence in the minibatch $n_{\max}$) of successively smaller suffixes starting at $i$, for each $i\in [1:n_{\max}]$ and use $\vectr{L}_s$ to compute the encodings $\vectr{h}_{s,i}$. The same procedure is repeated for the minibatch of sentences with tokens reversed  to compute $\vectl{h}_{s,i}$ and $\vectl{h}_{p,i}$.  As an optimization, several of the minibatches of the shorter suffixes can be combined to form larger minibatches. The worst case time complexity of computing all the representations is quadratic in  $n_{\max}$, as compared to the linear time complexity using a BiLSTM.  As we show in later sections, the increased time complexity is offset by the consistent gains in performance on several sentence modeling tasks. The encodings of the different can be computed in parallel, which can speed up computation greatly on modern hardware.

\section{Evaluation, Datasets, Training and Testing}
We evaluate the representational power of SuBiLSTM using several sentence modeling tasks and datasets from NLP. We do not concern ourselves with designing new models for SuBiLSTM. Rather, for each task, we take a strongly performing base model that uses the token representations of a BiLSTM and replace it with SuBiLSTM. The training procedures are kept exactly the same. 

\subsection{General Sentence Representation}
First, we investigate whether a SuBiLSTM can be trained to produce good general sentence representations that transfer well to several NLP tasks. As the base model, we use the recently proposed InferSent \cite{infersent}. It was shown to give strong results on a set of 10  NLP tasks encapsulated in the SentEval benchmark \cite{Conneau2018SentEvalAE}. The representation of a sentence is a max-pooling of the token representations produced by a SuBiLSTM. 
\begin{eqnarray}
\mathbf{H}^{\text{SuBiLSTM}}(\mathbf{s}) =  \max_{i\in [1:n]} \mathbf{H}_i^{\text{SuBiLSTM}}
\label{lab:maxsent}
\end{eqnarray}
where $\mathbf{H}_i^{\text{SuBiLSTM}}$ is defined in \eqref{eq:subidef}. The representation for $\mathbf{H}_i^{\text{SuBiLSTM-Tied}}$ is defined similarly. 
We train the model on the textual entailment task, where a pair of sentences (premise and hypothesis) needs to be classified into one of three classes - entailment, contradiction and neutral. 
Let $\mathbf{u}$ be the encoding of the premise according to \eqref{lab:maxsent} and let $\mathbf{v}$ be the encoding of the hypothesis. Using a Siamese architecture, the combined vector of $[\mathbf{u}; \mathbf{v}; |\mathbf{u}-\mathbf{v}|; \mathbf{u}\cdot \mathbf{v}]$ is used as the representation of the pair which is then passed through two fully connected layers and a final classification layer. 

\textbf{Training}. 
We use the combination of the Stanford Natural Language Inference (SNLI) \cite{Bowman2015} and the MultiNLI \cite{Williams2018} datasets to train. 
We set the hidden dimension of the LSTMs in SuBiLSTM to 2048, which produces a 4096 dimensional encoding for each sentence. The two fully connected layers are of 512 dimensions each.  The tokens in the sentence are embedded using GloVe embeddings \cite{Pennington2014} which are not updated during training.  We follow the same training procedure used for training the InferSent model in \cite{infersent}.


\textbf{Testing}. We test the sentence representations learned by SuBiLSTM on the SentEval benchmark.  This benchmark consists of 6 text classification tasks (MR, CR, SUBJ, MPQA, SST, TREC) with accuracy as the performance measure. There is one task on paraphrase detection (MRPC) with accuracy and F1  and one on entailment classification (SICK-E) with accuracy as the performance measure, respectively.  There are two tasks on textual semantic relatedness (SICK-R and STSB) for which the performance measure is the Pearson correlation between the predicted scores and ground truth. 
\begin{table}
  \centering
  \begin{tabular}{llll}
    \hline
     Dataset & Classification Task & \#Classes & Size  \\ \hline \hline
SST-2 & Sentiment  & 2  & 56.4k \\
SST-5 & Fine-grained Sentiment  & 5   & 94.2k \\
TREC-6 & Question  & 6  & 4.3k \\
TREC-50 & Fine-grained Question  & 50  & 4.3k \\
SNLI & Entailment  & 3  & 550k \\
MultiNLI & Entailment  & 3  & 393k \\
QUORA & Paraphrase  & 2  & 384k \\
\hline
   \end{tabular}
   \caption{Summary of the training datasets used in the evaluation of SuBiLSTM.}
   \label{tab:datasets}
\end{table} 
\begin{table*}
\begin{tabular}{lcccccccccc}
\toprule
\textbf{Model}&\textbf{MR}&\textbf{CR}& \textbf{SUBJ} &  \textbf{MPQA} & \textbf{SST} & \textbf{TREC} & \textbf{MRPC} & \textbf{SICK-R} & \textbf{SICK-E} &\textbf{STSB}\\ 
\midrule
\multicolumn{11}{ l}{\textit{Other Existing Methods}} \\
\midrule
FastSent+AE               & 71.8 & 76.7 & 88.8 & 81.5 & - & 80.4 &  71.2/79.1 & - &  - & -\\
SkipThought-LN  & {79.4} & {83.1} & {{93.7}} & {89.3} & {82.9} & {88.4} & -  &  {0.858} & 79.5 & -\\
DisSent & 80.1 & 84.9 &  93.6 &  90.1 & 84.1 &  93.6 & 75.0/- &  0.849 & 83.7 & - \\
CNN-LSTM            & 77.8 & 82.1 & 93.6 & 89.4 & -  & 92.6  & 76.5/83.8    &0.862  & -  & -\\
Byte mLSTM         & 86.9 & 91.4 & 94.6  & 88.5  & -   & -        & 75.0/82.8 & 0.792 & - & - \\
MultiTask       & 82.5 & 87.7 & 94.0 & 90.9 & 83.2 & 93.0 & 78.6/84.4 & 0.888 & 87.8 & 0.789 \\
QuickThoughts & 82.4 & 86.0 & 94.8 & 90.2 & 87.6 & 92.4 & 76.9/84.0 & 0.874 & - & - \\

\midrule
\multicolumn{11}{ l}{\textit{Supervised Training on AllNLI  (4096 dimensions)}} \\
\midrule
BiLSTM (InferSent) &  81.1 & 86.3 & 92.4 & 90.2 & 84.6 & 88.2 & 76.2/83.1 & 0.884 & 86.3  & 0.758\\
BiLSTM-2layer &  81.3 & 86.2 & 92.0 & 90.2 & \textbf{85.3} & 88.4 & 75.7/82.9 & 0.884 & 86.3  & 0.763\\
SuBiLSTM &  81.4 & 86.4 &  \textbf{93.2} &  \textbf{90.7} & 85.0 &  {89.8} &  \textbf{76.3/83.4} &  \textbf{0.886} & \textbf{86.7}  &  {0.770}\\
SuBiLSTM-Tied & \textbf{81.6} &  \textbf{86.5} & 93.0 & 90.5 &  {85.1} & \textbf{90.4} & 76.3/83.3 & 0.885 &  {86.3}  &  \textbf{0.771}\\
\bottomrule
\end{tabular}
%
\caption{Performance of SuBiLSTM on the SentEval benchmark. The first 8 methods contain both unsupervised and supervised ones. FastSent is from \cite{hill2016}, SkipThought is described in \cite{Kiros2015}, DisSent in \cite{Nie2017}, CNN-LSTM in \cite{Gan2017}, Byte mLSTM in \cite{Radford2017}, QuickThoughts in \cite{Logeswaran2018} and MultiTask in \cite{Sandeep2018}. Our base model is InferSent \cite{infersent}. Bold indicates the best performance among the SuBiLSTM models and the base model. }
\label{tab:senteval}
\end{table*}
\subsection{Text Classification}
We pick two representative tasks for text classification -- sentiment classification and question classification. As the base model, we use the Biattentive-Classification-Network (BCN) proposed by \cite{Cove2017}, which was shown to give strong performance on several text classification datasets, especially in association with CoVe embeddings \cite{Cove2017}. The BCN model uses two BiLSTMs to encode a sentence.
The intermediate states of the first BiLSTM are used to compute a self-attention matrix. This is followed by further processing and a second BiLSTM  before a final classification layer. Our hypothesis is that the richer contextual representations of SuBiLSTM should help such attention based sentence models. 
For our experiments, we replace only the first BiLSTM with a SuBiLSTM.

\textbf{Training and Testing}. For sentiment classification, we use the Stanford Sentiment Treebank dataset \cite{Socher2013RecursiveDM}, both in its binary (SST-2) and fine-grained (SST-5) forms. For question classification, we use the TREC \cite{Voorhees2001} dataset, both in its 6 class (TREC-6) and 50 class  (TREC-50) forms. The hidden dimension of the LSTMs  is set to 300. 
Distinct from \cite{Cove2017}, we use dropout after the embedding layer and before the classification layer. The two maxout layers are fixed at reduction factors of 4 and 2. We also apply weight decay to the parameters during optimization, which is done using Adam \cite{Kingma2015} with a learning rate of 1e-3. 
We experiment with two versions of the initial embedding -- one using GloVe only and the other using both GloVe and CoVe, both of which are fixed during training. 
Validation and testing are done using the sets associated with the SST and TREC datasets. 
\begin{figure}[t]
\begin{tikzpicture}
\usetikzlibrary{patterns}
\begin{axis}[ybar,
ymax=3,
bar width=4pt,
xtick={1, 2,3,4,5,6,7,8,9,10},
xticklabels={MR,CR,SUBJ,MPQA,SST,TREC,MRPC,SICK-R,SICK-E,STSB},
x tick label style={rotate=45},
ylabel={Percentage},
y label style={at={(0.07,0.5)}},
legend cell align={left},
legend entries={SuBiLSTM (Avg. $\Delta$ = 0.58\%), SuBiLSTM-Tied (Avg. $\Delta$ = 0.59\%)},
legend pos=north west,
]
\addplot
[draw=blue, fill=blue] 
coordinates
	{(1,0.3) (2,0.1) (3,0.8) (4,0.5) (5,0.4) (6,1.6) (7,0.3) (8,0.2) (9,0.4) (10,1.2)};

\addplot
[draw=purple,fill=purple] 
coordinates 
	{(1,0.5) (2,0.2) (3,0.6) (4,0.3) (5,0.5) (6,2.2) (7,0.2) (8,0.1) (9,0) (10,1.3)};
\end{axis}
\end{tikzpicture}
\caption{Gains  by using SuBiLSTM in the SentEval tasks. For MRPC we use F1 percentage, for SICK-R and STSB we use 100$\times$Pearson correlation and for the rest accuracy percentages. The Avg. $\Delta$ is the average of the 10 values. }
\label{fig:senteval}
\end{figure}
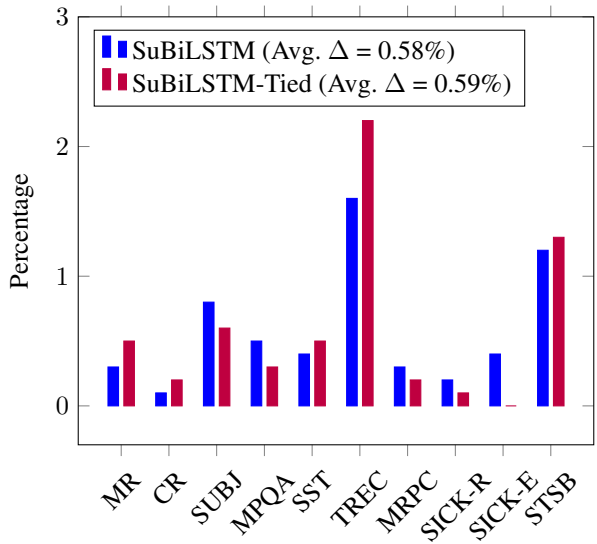
\subsection{Textual Entailment}
As mentioned above, the textual entailment problem is the task of classifying a pair of sentences into three classes -- entailment, contradiction and neutral. 
It is an important and canonical text matching problem in NLP.
To test SuBiLSTM for this task, we pick ESIM \cite{Chen2017} as the base model. ESIM has been shown to achieve state-of-the-art results on the SNLI dataset and has been the basis of further improvements. Like BCN above, ESIM uses two BiLSTM layers to encode sentences, with an inter-sentence attention mechanism in between. In our experiments, we only replace the first BiLSTM with a SuBiLSTM. 
\begin{table*}[t]
  \centering
  \begin{tabular}{llllll}
    \toprule
         & Model & Test & & Model & Test \\ \midrule
     
     \multirow{9}{*}{\STAB{\rotatebox[origin=c]{90}{{\textbf{SST-2}}}}}  
     & NSE \cite{Munkhdalai2017a}     & 89.7   & 
     \multirow{9}{*}{\STAB{\rotatebox[origin=c]{90}{{\textbf{TREC-6}}}}}  
     &   BCN+Char+CoVe \cite{Cove2017}  & 95.8   \\
         
     
     
     & BCN+Char+CoVe \cite{Cove2017}     & 90.3   & & TBCNN \cite{Mou2015}     & 96.0 \\
     
     & Byte mLSTM      & 91.8   & & LSTM-CNN \cite{Zhou2016}     & 96.1 \\
      & \cite{Radford2017}     &    & &    &\\
     \cline{2-3} \cline{5-6}    
      \addlinespace
     & BCN with BiLSTM   & 89.3   & & BCN with BiLSTM     & 95.2 \\
          & BCN with 2-layer BiLSTM   & 89.5   & & BCN with 2-layer BiLSTM     & 95.5 \\
         & BCN with SuBiLSTM             & 89.8   & &  BCN with SuBiLSTM      & 95.8 \\
     & BCN with SuBiLSTM-Tied       & 89.7  & & BCN with SuBiLSTM-Tied     & \textbf{96.2} \\
      \cline{2-3} \cline{5-6}   
      \addlinespace
          & BCN with BiLSTM+CoVe     & 90.1   & &  BCN with BiLSTM+CoVe      & 95.8 \\
             & BCN with 2-layer BiLSTM+CoVe     & 90.5   & &  BCN with 2-layer BiLSTM+CoVe      & 95.8 \\
           & BCN with SuBiLSTM+CoVe   & 91.0   & &  BCN with SuBiLSTM+CoVe     & 96.0 \\
     & BCN with SuBiLSTM-Tied+CoVe      & \textbf{91.2}   & & BCN with SuBiLSTM-Tied+CoVe     & 95.8 \\
\midrule

       \multirow{9}{*}{\STAB{\rotatebox[origin=c]{90}{{\textbf{SST-5}}}}}  
     & TE-LSTM \cite{Huang2017}     & 52.6  &
     \multirow{9}{*}{\STAB{\rotatebox[origin=c]{90}{{\textbf{TREC-50}}}}}  
     &  BCN+Char+CoVe \cite{Cove2017}   & 90.2   \\

     & NTI \cite{Munkhdalai2017b}     & 53.1   & & RulesUHC \cite{Silva2011}      & 90.8 \\
     
     &BCN+Char+CoVe \cite{Cove2017}  & 53.7 & & Rules \cite{Madabushi2016}     & 97.2 \\
     \cline{2-3} \cline{5-6}
     \addlinespace
     & BCN with BiLSTM   & 53.2   & & BCN with BiLSTM     & 89.8 \\
              & BCN with 2-layer BiLSTM             & 53.5   & &  BCN with 2-layer BiLSTM      & 89.4 \\
         & BCN with SuBiLSTM             & 53.2   & &  BCN with SuBiLSTM      & 89.8 \\
     & BCN with SuBiLSTM-Tied       & 53.4  & & BCN with SuBiLSTM-Tied     & {89.4} \\
        \cline{2-3} \cline{5-6}
     \addlinespace
          & BCN with BiLSTM+CoVe     & 53.6   & &  BCN with BiLSTM+CoVe      & 90.0 \\
                    & BCN with 2-layer BiLSTM+CoVe     & 54.0   & &  BCN with 2-layer BiLSTM+CoVe      & 89.2 \\
           & BCN with SuBiLSTM+CoVe   & 54.5   & &  BCN with SuBiLSTM+CoVe     & \textbf{90.2} \\
     & BCN with SuBiLSTM-Tied+CoVe      & \textbf{56.2}   & & BCN with SuBiLSTM-Tied+CoVe     & \textbf{90.2} \\
     \bottomrule
\end{tabular}
\caption{Comparison of text classification methods on the four datasets - SST-2, SST-5, TREC-6 and TREC-50. For each of them, we show accuracy numbers for BCN with SuBiLSTM and BCN with BiLSTM (base model), both with and without CoVe embeddings. The best performing ones among these is shown in bold. }
\label{tab:textclass}
\end{table*} 

\textbf{Training and Testing}. 
We use 300 dimensional GloVe embeddings to initialize the word embeddings (which are also updated during training) and use 300 dimensional LSTMs. We follow the same training procedure as \cite{Chen2017}.
Validation and  testing are on the corresponding sets in the SNLI dataset.

\subsection{Paraphrase Detection}
In this task, a pair of sentences need to be classified according to whether they are paraphrases of each other. To demonstrate the effectiveness of SuBiLSTM in a model that does not use any attention mechanism on the token representations, we use the same Siamese architecture used for training general sentence representations described above, except with one fully connected layer at the end followed by ReLU activation.

\textbf{Training and Testing}. 
We use 300 dimensional GloVe embeddings to initialize the word embeddings (which are also updated during training) and use 600 as the hidden dimension of all LSTMs and also the dimension of the fully connected layer. We apply dropout after the word embedding layer and after the ReLU activation. Training is done using the Adam optimizer with a learning rate of 1e-3. We use the QUORA dataset \cite{Quora} to train and test our models. 
A summary of the various datasets used in our evaluation is given in Table \ref{tab:datasets}. 
\subsection{Baselines}
For each of the tasks, we compare SuBiLSTM and SuBiLSTM-Tied with a single-layer BiLSTM and a 2-layer BiLSTM encoder with the same hidden dimension. While a SuBiLSTM-Tied encoder has the same number of parameters as single-layer BiLSTM, a SuBiLSTM has twice as many. In contrast, a 2-layer BiLSTM has more parameters than either of the SuBiLSTM variants if the hidden dimension is at least as large as the input dimension, which is the case in all out models. By comparing with a 2-layer BiLSTM baseline, we account for the larger number of parameters used in SuBiLSTM and also check whether the long range contextual information captured by SuBiLSTM can easily be replicated by adding more layers to the BiLSTM.

\section{Experimental Results}
In this section, for the sake of brevity, the terms SuBiLSTM and SuBiLSTM-Tied will sometimes refer to the base models where the BiLSTM has been replaced by our models. 

\subsection{General Sentence Representation}
The performance of SuBiLSTM and SuBiLSTM-Tied on the 10 transfer tasks in SentEval is shown in Table \ref{tab:senteval}. In all the tasks, SuBiLSTM and SuBiLSTM-Tied matches or exceeds the performance of the base model InferSent that uses a BiLSTM. For SuBiLSTM, among the classification tasks, the gains for SUBJ (0.8\%), MPQA (0.5\%) and TREC (1.6\%) over InferSent are particularly notable.  There is also a substantial gain of 1.2\% in the semantic textual similarity task (STSB). The performance of SuBiLSTM-Tied also follows a similar trend, gaining 0.6\% for SUBJ, 0.5\% for SST , 2.2\% for TREC and and 1.3\% for STSB.  The better performance on STSB is noteworthy as the sentence representations derived from a SuBiLSTM can take advantage of the long range dependencies it encodes.
The 2-layer BiLSTM based model performs comparably to  the single layer BiLSTM, despite using a much larger number of parameters. 

In Fig. \ref{fig:senteval} we plot the absolute gains made by SuBiLSTM and SuBiLSTM-Tied over BiLSTM for all the 10 tasks. It is interesting to note that both models perform comparably on an average, although SuBiLSTM has twice as many parameters as SuBiLSTM-Tied.
The performance of our models is still some way off from MultiTask \cite{Sandeep2018}; but they use a training dataset which is two orders of magnitude larger with a complex set of learning objectives. QuickThoughts \cite{Logeswaran2018} also uses a much larger unsupervised dataset. It is possible that SuBiLSTM coupled with training objectives and datasets used in these two works will provide substantial gains over the existing results. 


\subsection{Text Classification}
The performance of SuBiLSTM and SuBiLSTM-Tied on the four text classification datasets is shown in Table \ref{tab:textclass}. In three of these tasks (SST-2, SST-5 and TREC-50), SuBiLSTM-Tied using GloVe and CoVe embeddings performs the best. It performs notably better than the single layer BiLSTM based base model BCN on SST-2 and SST-5, achieving a new state-of-the-art accuracy of 56.2\% on fine-grained sentiment classification (SST-5). On TREC-6, the best result is obtained for SuBiLSTM-Tied using GloVe embeddings only, a new state-of-the-art accuracy of 96.2\%.There is no substantial improvement on the TREC-50 dataset. 

For text classification, we observe that SuBiLSTM-Tied performs better than SuBiLSTM  and CoVe embeddings give a  boost in most cases. The performance of the base model BCN with a 2-layer BiLSTM is slightly better than with the single layer BiLSTM in all cases except TREC-50. However, despite using a larger number of parameters, it does not perform better than both SuBiLSTM and SuBiLSTM-Tied. This implies that the richer contextual information captured by a SuBiLSTM cannot easily be replicated by adding more layers to the BiLSTM. Note that BCN uses a self-attention mechanism on top of the token representations and it is able to exploit the richer representations provided by SuBiLSTM.

\subsection{Textual Entailment}

\begin{table}[t]
  \centering
  \begin{tabular}{ll}
    \toprule
     Model & Test  \\ \midrule
     ESIM with BiLSTM \cite{Chen2017} & 88.0 \\
     DIIN \cite{Gong2018} & 88.0 \\
     BCN+Char+CoVe \cite{Cove2017} & 88.1 \\
     DR-BiLSTM \cite{Ghaeini2018} & 88.5 \\
     CAFE \cite{Tay2018} & 88.5 \\
     
     \midrule
     \midrule
     ESIM with BiLSTM (Ours) & 87.8 \\
     ESIM with 2-layer BiLSTM (Ours) & 87.9 \\
     ESIM with SuBiLSTM & \textbf{88.3} \\
     ESIM with SuBiLSTM-Tied & {88.2} \\
     \midrule
     ESIM with BiLSTM (Ensemble) & 88.6 \\
     ESIM with 2-layer BiLSTM (Ensemble) & 88.7 \\
     ESIM with SuBiLSTM (Ensemble) & \textbf{89.1} \\
     ESIM with SuBiLSTM-Tied (Ensemble) & \textbf{89.1}\\ 
          \bottomrule
\end{tabular}
\caption{ Accuracy of SuBiLSTM  and BiLSTM on the SNLI test set with ESIM as the base model.  }
\label{tab:snli_result}
\end{table} 

The performance of SuBiLSTM and SuBiLSTM-Tied on the SNLI dataset is shown in Table \ref{tab:snli_result}. Our implementation of ESIM, when using a BiLSTM, achieves 87.8\% accuracy. Using a SuBiLSTM, the accuracy jumps to 88.3\% and to 88.2\% for the Tied version. On using the 2-layer BiLSTM, accuracy improves only marginally by 0.1\%. This is aligned with the results shown for text classification above. 
Here again, the attention mechanism on top of the token representations benefit from the long range contextual information captured by SuBiLSTM. Note that ESIM uses an inter-sentence attention mechanism and is able to exploit the better token representations provided by SuBiLSTM across sentences. 
We also report the performance of an ensemble of 5 models. Both the SuBiLSTM versions achieve an accuracy of 89.1\%, while the BiLSTM based ones perform worse. 
\begin{table}[t]
  \centering
  \begin{tabular}{lll}
    \toprule
     Model  & Test \\
                \midrule
                BIMPM \cite{Wang2017BilateralMM}  & 88.2  \\
                pt-DECATTchar \cite{Tomar2017NeuralPI} &   88.4\\
                DIIN \cite{Gong2018} &  {89.1}\\
               MwAN \cite{Tan2018MultiwayAN} &   {89.1} \\
                \midrule
     \midrule
     BiLSTM & 87.8 \\
     2-layer BiLSTM & 87.9 \\
     SuBiLSTM & \textbf{88.2} \\
     SuBiLSTM-Tied & {88.1} \\
     \bottomrule
\end{tabular}%
\caption{Accuracy  of SuBiLSTM  and BiLSTM on the QUORA test set with a Siamese base model. All previous results use attention mechanisms. }
\label{tab:quora_results}
\end{table} 
\begin{figure}[t]
\begin{tikzpicture}
\usetikzlibrary{patterns}
\begin{axis}[ybar,
ymax=2.7,
bar width=4pt,
xtick={1, 2,3,4,5,6,7,8,9,10},
xticklabels={SentEval, SST-2, SST-5, TREC-6, TREC-50, SNLI, QUORA},
x tick label style={rotate=45},
ylabel={Percentage},
y label style={at={(0.06,0.5)}},
legend cell align={left},
legend entries={SuBiLSTM , SuBiLSTM-Tied },
legend pos=north east,
]
\addplot
[draw=blue, fill=blue] 
coordinates
	{(1,0.58) (2,0.9) (3,0.9) (4,0.6) (5,0.2) (6,0.5) (7,0.4)};

\addplot
[draw=purple,fill=purple] 
coordinates 
	{(1,0.59) (2,1.1) (3,2.6) (4,1.0) (5,0.2) (6,0.4) (7,0.3)};
\end{axis}
\end{tikzpicture}
\caption{Gains from using SuBiLSTM and SuBilSTM-Tied over single layer BiLSTM on all the datasets. The difference is between the best figures obtained for each model. For SentEval we use the average score and accuracy for the rest.}
\label{fig:subicompare}
\end{figure}
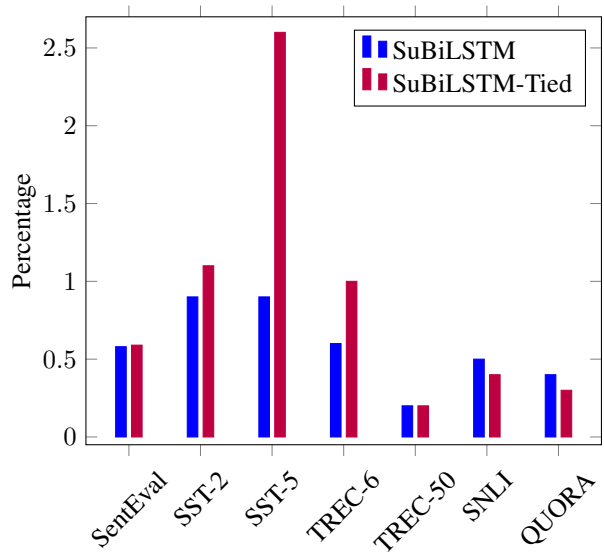

\subsection{Paraphrase Detection}
The accuracies obtained on the QUORA dataset are shown in Table \ref{tab:quora_results}. Note that unlike the BCN and ESIM models, we use a simple Siamese architecture without any attention mechanism. In fact, the representation of a sentence in this case is simply the max-pooling of all the intermediate representations of the SuBiLSTM. Even in this case, we observe gains over both single layer and 2-layer BiLSTMs, although slightly lesser than the attention based models. The best model (SuBiLSTM) achieves 88.2\%, at par with a more complex attention based model BiMPM \cite{Wang2017BilateralMM}.

\subsection{Comparison of SuBiLSTM and SuBiLSTM-Tied}
The results shown above clearly show the efficacy of using SuBiLSTMs in existing models geared towards four different  sentence modeling tasks.  The relative performance of SuBiLSTM and SuBiLSTM-Tied are fairly close to each other, as shown by the relative gains in Fig. \ref{fig:subicompare}. SuBiLSTM-Tied  works  better on small datasets (SST and TREC), probably owing to the regularizing effect of using the same LSTM to encode both suffixes and prefixes. For the larger datasets (SNLI and QUORA), SuBILSTM slightly edges out the tied version owing to its larger capacity. The training complexity for both the models is similar and hence, with half the parameters, SuBILSTM-Tied should be the more favored model for sentence modeling tasks. 

\section{Related Work}
Recurrent Neural Networks \cite{Elman1990} have emerged as one of the most powerful tools for computing distributed representations of sequential data. The problems of training vanilla RNNs \cite{Bengio1994} were addressed by more sophisticated models -- most notably the Long Short Term Memory (LSTM)  \cite{Hochreiter1997} and the simpler GRU \cite{Cho2014}. Over the years, several alternatives to the basic RNN model have been proposed. A Dilated-RNN \cite{Chang2017} uses progressively dilated connections between recurrent nodes to extract long range dependencies efficiently. A Skip-RNN \cite{Chang2017} learns to skip state updates rather than applying them at each token in a sequence, thereby achieving faster training and inference times.  Recurrent Highway Networks \cite{Zilly2017} allow for multiple state updates via highway connections at each time step and a Clockwork-RNN \cite{Koutnik2014} updates its state at multiple timescales.  The idea of capturing long term dependencies in better ways has given rise to memory augmented architectures like Neural Turing Machines \cite{Graves2014} and TopicRNNs \cite{Dieng2017}. 

In this paper, we focus on LSTMs. As shown by the work of \cite{Jozefowicz2015} and \cite{Greff2016}, LSTMs represent a robust recurrent neural network architecture for modeling sequential data. In particular, LSTMs are a core component in several state-of-the-art neural models for NLP tasks like language modeling \cite{Melis2018,Merity2018}, textual entailment \cite{Chen2017}, question answering \cite{Seo2017}, semantic role labeling \cite{He2017} and named entity recognition \cite{MaHovy2016}. 

A unidirectional RNN processes a sequence in a single direction, usually following the natural  order specific to the sequence. Bidirectional RNNs, where two distinct recurrent networks process the input sequence in opposite directions was first proposed by \cite{Schuster1997}. This allows the model to have a representation of the prefix and the suffix at each intermediate point in the sequence, thereby providing context in both directions. Following the work by \cite{Graves2005}, Bidirectional LSTMs have become a mainstay for sequence representation tasks. The concept of having encodings of different contexts has since been generalized to Multidimensional LSTMs \cite{Graves2008} and Grid LSTMs \cite{Kalchbrenner2016}. 

In the recently proposed Twin-Networks \cite{Serdyuk2018}, the authors show that forcing the prefix encoding in a BiLSTM to be close to the suffix encoding in the reverse direction acts as a regularizer and helps capture more long term dependencies. We take a more direct approach -- explicitly encoding the suffix in the forward direction and forcing an interaction with the prefix encoding through a max-pooling.  
Although we focus on LSTMs in this paper, our idea generalizes trivially to other RNN cells.  

\section{Conclusion}
We propose SuBiLSTM and SuBiLSTM-Tied, a simple, general and effective improvement to the BiLSTM model, where the prefix and suffix of each token in a sentence is encoded in both forward and reverse directions to capture long range dependencies. We demonstrate gains in performance by replacing BiLSTMs in existing models for several sentence modeling tasks.
The main drawback of our method is the quadratic time complexity required to compute the representations in a SuBiLSTM. As future direction of work, we intend to explore variants of  SuBiLSTM, where only suffixes of fixed  or small random lengths are computed. We also plan to utilize the information (e.g. encodings of subsequences) exposed by SuBiLSTM in more novel ways.

\bibliographystyle{aaai}
\bibliography{bibtex}

\end{document}